# Developing and validating multi-modal models for mortality prediction in COVID-19 patients: a multi-center retrospective study

Joy Tzung-yu Wu, MBBS[1,*], Miguel Ángel Armengol de la Hoz, PhD[2,3,4,*,†], Po-Chih Kuo, PhD[2,5,*,†], Joseph Alexander Paguio, MD[6,9], Jasper Seth Yao, MD[6,9], Edward Christopher Dee, MD[7], Wesley Yeung, MBBS[2,8], Jerry Jurado, MD[9], Achintya Moulick, MD[9], Carmelo Milazzo, MD[9], Paloma Peinado, MD[10], Paula Villares, MD[10], Antonio Cubillo, MD[10], José Felipe Varona, MD[10], Hyung-Chul Lee, MD[11], Alberto Estirado, BS[10], José Maria Castellano, MD[10,12,#], Leo Anthony Celi, MD[2,13,14,#]

[1] Department of Radiology and Nuclear Medicine, Stanford University, Palo Alto, CA, United States.

[2] Institute for Medical Engineering and Science, Massachusetts Institute of Technology, Cambridge, MA, United States.

[3] Department of Anesthesia, Critical Care and Pain Medicine, Beth Israel Deaconess Medical Center, Boston, MA, United States

[4] Big Data Department, Fundacion Progreso y Salud, Regional Ministry of Health of Andalucia, Andalucia, Spain.

[5] Department of Computer Science, National Tsing Hua University, Hsinchu, Taiwan.

[6] Albert Einstein Medical Center, Philadelphia, PA, United States.

[7] Department of Radiation Oncology, Memorial Sloan Kettering Cancer Center, New York, NY, United States.

[8] National University Heart Center, National University Hospital, Singapore, Singapore.

[9] Hoboken University Medical Center–CarePoint Health, Hoboken, NJ, United States.

[10] Centro Integral de Enfermedades Cardiovasculares, Hospital Universitario Monteprincipe, Grupo HM Hospitales, Madrid, Spain.

[11] Department of Anesthesiology and Pain Medicine, Seoul National University College of Medicine, Seoul, Republic of Korea.

[12] Centro Nacional de Investigaciones Cardiovasculares, Instituto de Salud Carlos III, Madrid, Spain.

[13] Department of Medicine, Beth Israel Deaconess Medical Center, Boston, MA, United States.

[14] Department of Biostatistics, Harvard T.H. Chan School of Public Health, Boston, MA, United States.

*co-first authors

#co-senior authors

†Correspondence to:
Po-Chih Kuo
Address: No. 101, Section 2, Guangfu Rd, East District, Hsinchu City, Taiwan
Email: kuopc@cs.nthu.edu.tw
Telephone: +886342973

Miguel Ángel Armengol de la Hoz

Email: maarmeng@mit.edu

The work was originated from Institute for Medical Engineering and Science, Massachusetts Institute of Technology, Cambridge, MA, United States.

Manuscript Type: Original Research

Word Count for Text: 3399

# Abstract

**Background**

The unprecedented global crisis brought about by the COVID-19 pandemic has sparked numerous efforts to create predictive models for the detection and prognostication of SARS-CoV-2 infections with the goal of helping health systems allocate resources. Machine learning models, in particular, hold promise for their ability to leverage patient clinical information and medical images for prediction. However, most of the published COVID-19 prediction models thus far have little clinical utility due to methodological flaws and lack of appropriate validation.

**Purpose**

In this paper, we describe our methodology to develop and validate multi-modal models for COVID-19 mortality prediction using multi-center patient data.

**Materials and Methods**

The models for COVID-19 mortality prediction were developed using retrospective data from Madrid, Spain (N=2547) and were externally validated in patient cohorts from a community hospital in New Jersey, USA (N=242) and an academic center in Seoul, Republic of Korea (N=336). The models we developed performed differently across various clinical settings, underscoring the need for a guided strategy when employing machine learning for clinical decision-making.

**Results**

We demonstrated that using features from both the structured electronic health records and chest X-ray imaging data resulted in better 30-day-mortality prediction performance across all three datasets (areas under the receiver operating characteristic curves: 0.85 (95% confidence interval: 0.83 – 0.87), 0.76 (0.70 – 0.82), and 0.95 (0.92 – 0.98)). We discuss the rationale for the decisions made at every step in developing the models and have made our code available to the research community.

**Conclusion**

We employed the best machine learning practices for clinical model development. Our goal is to create a toolkit that would assist investigators and organizations in building multi-modal models for prediction, classification and/or optimization.

# Introduction

Beginning as an outbreak of an unknown viral pneumonia in Wuhan, China, the coronavirus disease 2019 (COVID-19) pandemic has sparked numerous efforts to create predictive models. In particular, machine learning methods hold great promise because they provide the opportunity to combine and use features from multiple modalities available in electronic health records (EHR), such as imaging and structured clinical data, for downstream prediction tasks. At present, there are hundreds of papers in preprint servers and medical journals employing machine learning methodologies in an attempt to bridge the gaps in the diagnosis, triage, and management of COVID-19; eight of them have integrated both radiological and clinical data.[1–8]

However, most of these studies were found to have little clinical utility, producing a credibility crisis in the realm of artificial intelligence in healthcare. A recent review by Roberts et al. found that, after screening more than 400 machine learning models using various risk and bias assessment tools, none of the evaluated machine learning models had sufficiently fulfilled all of the following: (1) documentation of reproducible methods, (2) adherence to best practices

in the development of a model, and (3) external validation that could justify claims of applicability.[9]

Furthermore, the question remains as to how useful these predictive models actually are to other institutions to which these models were not customized.[10] While machine learning models offer the potential for a more accurate prediction of clinical outcomes within a specific context, these models were usually trained using data from a single institution and are unable to identify differences in contexts when employed in other settings.[11] This problem raises the need for validation not just in the neighboring center, but in other types of centers, states, or even countries, where patient demographics, standards of care, institutional policies may largely differ. In addition, these models need to be constantly updated because the contexts in which these models were trained and approved for use may be significantly different when used at present day.[12–15] Finally, beyond concerns about the reproducibility and generalizability of machine learning models is the issue of the lack of explainability, in which models may draw spurious associations between confounding imaging features and the outcome of interest.[12,16] DeGrave et al. attempted to assess the trustworthiness of recently published machine learning models for COVID-19 by using explainable AI technology to determine which regions of chest X-rays (CXR) these models used to predict outcomes.[12] Surprisingly, they found that in addition to highlighting lung regions, the evaluated models used laterality marks, CXR text markers, and other features that provide no pathologic basis for distinguishing between COVID-positive and COVID-negative studies.[12] In other words, it was discovered that these models used shortcuts, further underscoring concerns about their applicability.

Therefore, we believe that some of the ways investigators can address the questions surrounding the credibility of a machine learning model are (1) to state the clinical context, which include patient demographics, geography, and timeframe, of the training and testing datasets that were used (2) to provide the resources for other centers to create or fine-tune models specific to their contexts, (3) to be explicit about the appropriate level of the model's generalizability based on results of external validation studies, (4) to explore strategies that either build in and/or evaluate the explainability of models, and (5) to externally validate the performance of the model on different subpopulations of the sample and explore the fairness of the model in underrepresented patient groups.

In our case, we present our efforts to develop three machine learning models for predicting 30-day mortality among hospitalized patients with COVID-19: (1) a structured EHR-based model, (2) a CXR-based model, and (3) an EHR-CXR fusion model. All three models were developed using a multi-center dataset from Madrid, Spain. We aim to investigate how the performance of each of these models differed when validated on two external unseen single-center datasets from different countries (U.S.A. and Republic of Korea). In addition, we will flag and detail why certain modeling design decisions were made, including the difficulties and trade-offs of these decision-making processes. The Checklist for Artificial Intelligence in Medical Imaging[17] is used to report our study designs and findings. We have made the code and other resources to reproduce our model training process available to the research community. We hope our work would serve as a toolkit that future investigators could use, adapt and retrain models using data from their own institutions. Ultimately, our goal is to provide other institutions the opportunity to leverage machine learning technology to predict the mortality of their patients with COVID-19 and customize these models to meet their individual institution's needs.

# Methods

**Study objectives**

We used retrospective data from Hospitales de Madrid to build three machine learning models that taking input from 1) only structured EHR data, 2) the first CXR image, and 3) both the EHR data and CXR image for predicting COVID-19 patient's mortality at 30 days from hospital admission, as illustrated in Figure 1. Our aims are: 1) To investigate if modeling with features from both EHR and CXR image data result in improved mortality prediction performance. 2) To investigate if modeling with features from both EHR and CXR image data result in more consistent model performance on external test sets (from Hoboken, New Jersey, U.S.A. and Seoul, Republic of Korea).

**Datasets**

Three datasets of patients with confirmed SARS-CoV-2 infection from three different countries were used in our study (see details under Case definition). The hospital mortality outcomes came from the source EHRs.[18,19] Table 1 documents the number of patients included and/or excluded for different reasons. We used the dataset from Hospitales de Madrid (HM), a network of 17 hospitals in Madrid, Spain, for all model training and hyperparameter tuning experiments. The dataset is accessible via credentialed and HIPAA-compliant approvals from https://www.hmhospitales.com/coronavirus/covid-data-save-lives/english-version. External datasets from Hoboken University Medical Center (HUMC), USA and Seoul National University Hospital (SNUH), Republic of Korea were used for validating and evaluating the trained models. Table 2 describes the basic clinical characteristics of patients in each of the three different datasets. For more detailed patient clinical characteristics, please see Supplementary Table 1. Case Definitions and Ethical Statements are described in the supplementary materials.

**Data preprocessing**

We included both comorbidities and lab EHR variables for the EHR-based and the EHR-CXR fusion models. Variables (categorical) included for comorbidities are: diabetes, hyperlipidemia (HLD), hypertension (HTN), ischemic heart disease (IHD), chronic kidney disease, chronic obstructive pulmonary disease (COPD), asthma, cancer, chronic liver disease, stroke, congestive heart failure (CHF), and dementia. Lab variables (numeric) include: LDH, hemoglobin, MCV, neutrophil percentage, mean neutrophil, lymphocyte percentage, mean lymphocyte, mean leukocyte, mean platelet volume, mean platelet, CRP, MCH, AST, ALT, APTT, D-dimer, prothrombin activity, INR, glucose, sodium, potassium, BUN, and creatinine (See supplementary materials for the details of EHR data preprocessing). Images were included and excluded as per Table 1.

**Training and model picking**

Four different types of machine learning were tried in a tuning setting to select for the best EHR-based model. The CXR-based model building including 5 steps: 1) Online (real-time) image augmentation during training. 2) Online CXR feature extraction. 3) Mortality classification layers 4) Optimization settings 5) Hyperparameter tuning and model selection, as described in supplementary. For the fusion model, we took a late fusion approach that uses the output probability from the CXR model as a feature along with the EHR features for the 30-day mortality classification.

**Statistical analysis**

In all three datasets, we reported the point estimates and 95% confidence interval (95% CI) of the reported validation metrics: areas under the receiver operating characteristic curves (AUROC), sensitivity, specificity, positive predictive value (PPV), negative predictive value

(NPV), F1-score, and accuracy. 95% CIs were computed by bootstrapping the scores of the predictions 1000 times.

**Model evaluation**

Internal validation was performed by averaging the results across the 4 folds using the best hyperparameters for the final models using the Madrid dataset. External validation using the Hoboken and Seoul datasets were respectively performed by two in-house Hoboken physicians and a data scientist, all of whom were uninvolved in model tuning.

Specifically, validation of the Hoboken dataset was performed by clinicians of the community hospital (JAP and JSY) who had the appropriate credentials to access patient health information. To accomplish this task, the necessary code was developed by an external team of data scientists. This code is made available (Supplementary Table 6) to allow future researchers to replicate our methodology that allows inter-institutional collaboration while complying with data governance standards and protecting sensitive patient information.

**Evaluating model fairness**

A more complete analysis of biases in our models is not within the scope of our study, particularly since the datasets came from countries with completely different ethnic makeup. However, as a baseline, performance results for male and female are reported separately on all three datasets to explore how the models' performance differ between the gender strata.

**Evaluating model explainability**

We used SHapley Additive exPlanations (SHAP)[20] to show the feature importance in our EHR-based and fusion model.[20] The SHAP method estimates differences between models with different feature subsets and calculates SHAP values representing the importance of each feature to overall model predictions. The features with larger absolute SHAP values are supposed to contribute more to the prediction. A more positive SHAP value for a feature corresponds to a higher model predicted likelihood. For the fusion model, the SHAP analysis helps to show whether the CXR model's prediction is important for the final morality prediction. For evaluating the explainability of the CXR model's prediction, we visualize where on the image the model attended to most by using Grad-CAM.[21] Grad-CAM computes the gradient of the prediction scores of the features generated by convolutional layers to reveal which locations in the image are most important.

**Design decisions and reasons**

We listed all the decisions and reasons of the methodology in the supplementary materials.

**Data sharing**

We have made the code and other resources to reproduce our model training process available to the research community. The associated datasets in this study can be accessed through the respective application processes of the hospitals involved.

# Results

**Internal validation**

Table 3 shows the results of internal validation for mortality prediction in COVID-19 patients. The models were trained and validated on the Madrid dataset using 4-fold cross validation. The respective F1-score (95% Confidence interval) of the EHR-based, CXR-based, and fusion models were 0.36 (95% CI 0.32-0.41), 0.37 (0.33-0.41), and 0.40 (0.36-0.45). Figure 3 (A) shows the ROC curves obtained from EHR-based (orange), CXR-based (green), and fusion models (blue) for internal validation on Madrid datasets.

**External testing**

Table 4 shows the results of external testing for mortality prediction using all the Madrid dataset for model development and Hoboken and Seoul datasets for external testing. In the external testing on the Hoboken dataset, the F1-score (95% CI) of the EHR-based, CXR-based, and fusion models were 0.66 (0.59-0.73), 0.64 (0.57-0.70), and 0.69 (0.62-0.76), respectively. In the external testing on the Seoul dataset, the respective F1-score (95%CI) of the EHR-based, CXR-based, and fusion models were 0.15(0.04–0.28), 0.13 (0.03-0.25), and 0.21 (0.06-0.38). The ROC curves for external testing on Hoboken and Seoul datasets are illustrated in Figure 3 (B) and (C), respectively.

**Explainability analysis**

Figure 4 and 5 show the impact of features on the EHR-based and fusion models' prediction, respectively. Figure 5 shows the CXR model's prediction and the patient age were two features with highest mean absolute SHapley Additive exPlanations (SHAP)[20] value for predicting 30-day mortality. Figure 6 shows the mean Gradient-weighted Class Activation Mapping (Grad-CAM)[21] heatmap obtained by averaging the heatmaps with prediction probability larger than 0.6 for the expired patients in the Madrid dataset. CXR regions with high levels of importance in the model prediction (represented in red or yellow) were located within the lung zones. In particular, the lung parenchyma and mediastinal structures were the primary focus of the algorithm for mortality prediction.

**Fairness analysis**

Supplementary Table 5 shows the difference of model performance between female vs male patients across all three datasets. All the expired cases in the Seoul dataset are male.

# Discussion

Our findings demonstrate differences in the performance of our predictive models across different institutions, clinical settings, and populations. Previous studies have demonstrated that predictive models tend not to perform well outside the institution and setting that it was trained in, while also losing their accuracy over time due to underlying clinical data drift.[22–24] In this paper, we trained three models (EHR-based, CXR-based and fusion) by optimizing their F1-scores for 30-day-mortality prediction from hospital admission for confirmed COVID-19 patients. On internal validation (Madrid) and testing on two external datasets (Hoboken and Seoul), point estimates of the F1-score of the fusion model consistently outperformed the EHR-based and the CXR-based models. These findings are not statistically significant at 95% CI, which can be expected in the context of small numbers of mortality events in all the datasets. We reported all metrics for transparent reporting on our models' performance. Reporting just AUROC and/or accuracy can give a falsely higher sense of model performance for imbalanced datasets.

On evaluating the changes in F1-scores between the results on the Madrid dataset and the two external test sets, we see expected significantly drops in F1-scores for all three models when they are tested on the Seoul dataset (statistically significant). However, we see unexpectedly higher F1-scores from all three models for the Hoboken dataset (statistically significant). Possible explanations for these findings include 1) the surge in COVID-19 cases during this period in the Greater New York Metropolitan area, which includes Hoboken, NJ, that led to higher mortality rates in the hospital (more likely), and 2) the pre-trained 'teacher' CXR models used in the development of our CXR model were trained on CXR images from the United States (possible). Both factors would make Hoboken an easier evaluation dataset than the both the Madrid and Seoul datasets. For the former factor, F1-scores would naturally be higher if the target prevalence is higher in a dataset. For the latter factor, since deep learning models are brittle to small changes in machine type and calibration, changes in geography (a different country with likely more different machines and calibration protocols) alone may affect model's feature extraction suitability for the same task -- i.e. it is possible that the CXR models extract better features from Hoboken CXRs as the images were taken in similar US setting.

Our findings further highlight the need for a guided strategy with the use of predictive models as clinical decision support tools. This is particularly important for machine learning models because, unlike traditional statistically based models (e.g., multivariate regression models which have deterministic performance on the same data), the state-of-the-art machine learning models, though powerful, are often built with many design decisions during their development. Changes in any parameters, optimizing target(s) (e.g., accuracy, AUROC, and F1-score), or even just different GPUs or TPUs for training, could result in different performance. As such, performance reported in any paper is really a snapshot. Furthermore, institutional and temporal data variation can also change the performance of these models. Ultimately, performance of predictive models, like most medical tests, are sensitive to changes in disease severity, prevalence and distribution in a patient population.

Therefore, we reinforce the previously reported recommendation that institutions should reassess models on local datasets.[15,25] They should also consider fine tuning their own predictive models and to learn what works best for their local patient populations. Training a predictive model to prioritize a high positive predictive value may mitigate the risk of incorrectly predicting an outcome—in this case, mortality—among patients who would have otherwise survived or have not met the outcome. Prioritization of sensitivity may allow clinicians to triage as many severely ill patients to advanced care facilities. Ultimately, locally tested predictive models can become tools that help institutions address their individual needs, either to appropriately distribute limited resources or to help detect and manage severely ill patients early. To facilitate this and for reproducibility, we have open sourced our data preprocessing and training code for re-use by other research groups and institutions.

Multiple groups have suggested best practices and regulations in the development and use of artificial intelligence and predictive modeling that address pertinent concerns.[15,25] Much of these recommendations have to do with reproducibility, quality of data being used, and the intended function of artificial intelligence programs.[33] However, the complexity of applying machine learning models in clinical settings goes beyond reproducibility and generalizability. Deep learning's advantage of not needing to engineer predictive features for model building can be offset by the disadvantage of its lack of explainability. Models may draw spurious associations between confounding tabular or imaging features and the outcome of interest.[12,16] For example, a prior study has shown that AI has a predilection for detecting imaging features other than signals of pathology as shortcuts for predicting COVID-19 outcomes.[12] Unlike linear models, weights in deep learning models have no intrinsic significance on their own that can be interpreted clinically outside the model. Despite this, in the imaging space, researchers have used methods, such as Grad-CAM, to post analyze the 'explainability' of deep learning imaging models by examining where on the image the trained model 'attended' to most for prediction. However, these analyses are often qualitative in nature for publication purposes, which we argue is insufficiently rigorous for most clinical applications.

Therefore, for imaging feature explainability, we not only presented heatmaps from post-training Grad-CAM analysis of the CXR model, but also specifically used anatomical bounding box augmentation to teach our CXR model to focus on lungs and mediastinum regions for prediction during the training stage. Choices for both tuning with or without anatomy augmentation was used and the best model had utilized the anatomical regional augmentation approach. As shown in Grad-CAM analysis, our CXR model does focus on the clinically important lungs and mediastinum CXR regions for prediction. This gives our clinicians more confidence for the model's prediction.

Lastly, we took a late fusion approach to model features from the EHR data and the CXR image so that we could easily assess how much the prediction depended on different data sources via SHAP plots. Previous machine learning models have also used similar techniques to elucidate the explainability of their models.[12,28] In our case, the fusion model placed the most weight on patient age and CXR features, further supporting that including both clinical and imaging data improves downstream model performance.

This study is limited by its retrospective nature, imbalanced and small evaluation datasets, and merits further research. The models trained also need the exact same input features at inference and some institutions may not routinely collect all the required input variables. In addition, as with many published AI models, fairness as an operationalized outcome has not been incorporated in our models.[25] We did, however, assess for differences in the models' performance separately for male and female on the Madrid and the Hoboken dataset (no female deaths in the Seoul dataset), which showed no statistically significant differences on 95% CI analysis except the EHR-based model on the Madrid dataset. In general, evaluating differences in model performance in subpopulations can help elucidate and inform downstream applications about potential problems if an AI model was applied to patients from under-represented/marginalized populations. Pooled results from the general population may gloss over worse outcomes in vulnerable groups.[29] Furthermore, in general when the dataset contains underlying data entry biases and/or imbalanced representations, building fair models is still an unsolved technical research problem.

Machining learning methods offer the advantage of utilizing richer clinical data for predictive modeling, which many have explored during the COVID-19 pandemic. However, many studies published so far have further exposed the credibility crisis that machine learning is facing in terms of reproducibility, generalizability, explainability and fairness. This is often due to implementation issues, such as poorly documented study designs, lack of external test sets and study code availability. In this paper, we employed best machine learning practices and trained three machine learning models on a model development (internal) dataset. We subsequently stress tested the final models on two external datasets from different countries. We redemonstrated 1) that using features from both the EHR and CXR imaging data resulted in better 30-day-mortality prediction performance across all three datasets, and 2) the need to fine tune models on local datasets and update with time. We evaluated our models for explainability in terms of feature dependence, and fairness in terms of gender-based performance differences. Finally, for the sake of transparency and reproducibility, we documented all study design decisions and made the study code available to the research community.

# Tables

**Table 1:** High level descriptive summary of datasets used in this study.

| Dataset name | Data split | Inclusion criteria | Exclusion criteria | | Size (number included / all)* |
|---|---|---|---|---|---|
| **Madrid** | 4-fold training and internal validation for building and tuning the models | Multicentered hospital network, Madrid, Spain, from 12/2019-06/2020[18] | Under age (< 16) | N = 14 | 1628 / 2547 |
| | | | Missing admission time | N = 85 | |
| | | | Missing admission chest X-Ray. | N = 820 | |
| **Hoboken** | Test (External validation) | Community hospital, Hoboken, NJ, USA, from 03/2020-04/2020[19] | Under age (< 16) | N = 0 | 201 / 242 |
| | | | Missing admission time | N = 0 | |
| | | | Missing admission chest X-ray | N = 41 | |
| **Seoul** | Test (External validation) | Academic tertiary hospital, Seoul, Republic of Korea, from 1/1/2020-12/31/2020 | Under age (< 16) | N = 16 | 315 / 336 |
| | | | Missing admission time | N = 0 | |
| | | | Missing admission chest X-ray | N = 5 | |

Note: *These are unique patients.

**Table 2:** Summary of clinical characteristics for the 3 different datasets used in the study.

| Characteristics | Madrid | | | Hoboken | | | Seoul | | |
|---|---|---|---|---|---|---|---|---|---|
| | Alive | Expired | p value | Alive | Expired | p value | Alive | Expired | p value |
| n | 1439 | 189 | | 114 | 87 | | 310 | 5 | |
| Age (mean) | 65.7 | 79.6 | <0.001 | 61.9 | 69.1 | 0.003 | 45.7 | 64.0 | 0.053 |
| Female (%) | 41.3 | 28.6 | <0.001 | 48.2 | 32.2 | 0.032 | 48.4 | 0 | 0.062 |
| 30-day mortality (%) | 88.4 | 11.6 | <0.001 | 56.7 | 43.3 | <0.001 | 98.4 | 1.6 | <0.001 |
| Diabetes (%) | 16.1 | 24.1 | 0.008 | 39.4 | 35.6 | 0.682 | 11.0 | 40.0 | 0.103 |
| Hypertension (%) | 6.1 | 10.2 | 0.055 | 54.3 | 55.2 | 0.974 | 13.9 | 80.0 | 0.002 |
| Hyperlipidemia (%) | 26.0 | 35.3 | 0.01 | 32.5 | 34.5 | 0.880 | 6.1 | 20.0 | 0.283 |
| Congestive Heart Failure (%) | 4.3 | 7.0 | 0.156 | 16.7 | 14.9 | 0.891 | 1.6 | 20.0 | 0.093 |
| Ischemic Heart Disease (%) | 6.0 | 13.4 | <0.001 | 16.7 | 14.9 | 0.891 | 2.3 | 20.0 | 0.122 |
| Stroke (%) | 2.6 | 7.5 | <0.001 | 2.6 | 4.6 | 0.469 | 2.6 | 0.0 | - |
| COPD (%) | 4.7 | 8.0 | 0.083 | 9.6 | 10.3 | 0.941 | 0.6 | 20.0 | 0.047 |
| CKD (%) | 5.1 | 11.8 | <0.001 | 7.0 | 20.7 | 0.008 | 1.3 | 20.0 | 0.078 |
| Chronic Liver Disease (%) | 0.6 | 4.8 | <0.001 | 0.9 | 0.0 | 1.000 | 1.0 | 0.0 | - |
| Active Cancer (%) | 4.0 | 12.3 | <0.001 | 6.1 | 3.4 | 0.519 | 0.0 | 0.0 | - |

Note: COPD: Chronic Obstructive Pulmonary Disease; CKD: Chronic Kidney Disease

**Table 3.** Internal validation on Madrid dataset with 95% confidence intervals

|  | EHR-based | CXR-based | Fusion |
|---|---|---|---|
| **AUROC (CI)** | 0.82 (0.79-0.84) | 0.81 (0.78-0.83) | 0.85 (0.83-0.87) |
| **Sensitivity (CI)** | 0.77 (0.71-0.82) | 0.76 (0.71-0.82) | 0.79 (0.74-0.84) |
| **Specificity (CI)** | 0.71 (0.66-0.76) | 0.72 (0.67-0.75) | 0.74 (0.71-0.78) |
| **PPV (CI)** | 0.24 (0.21-0.28) | 0.25 (0.21-0.28) | 0.27 (0.23-0.31) |
| **NPV (CI)** | 0.96 (0.95-0.97) | 0.96 (0.95-0.97) | 0.97 (0.96-0.98) |
| **F1-score (CI)** | 0.36 (0.32-0.41) | 0.37 (0.33-0.41) | 0.40 (0.36-0.45) |
| **Accuracy (CI)** | 0.71 (0.68-0.76) | 0.73 (0.68-0.76) | 0.75 (0.72-0.78) |

Note: AUROC: area under the receiver operating characteristic curve; PPV: positive predictive value; NPV: negative predictive value; CI: confidence interval.

**Table 4.** External testing on Hoboken and Seoul datasets with 95% confidence intervals

|  | Hoboken dataset | | | Seoul dataset | | |
|---|---|---|---|---|---|---|
|  | EHR-based | CXR-based | Fusion | EHR-based | CXR-based | Fusion |
| **AUROC (CI)** | 0.74 (0.68-0.80) | 0.72 (0.66-0.78) | 0.76 (0.70-0.82) | 0.92 (0.88-0.96) | 0.90 (0.86-0.94) | 0.95 (0.92-0.98) |
| **Sensitivity (CI)** | 0.68 (0.59-0.77) | 0.68 (0.57-0.8) | 0.68 (0.60-0.76) | 0.64 (0.25-0.86) | 0.63 (0.20-0.86) | 0.64 (0.20-0.86) |
| **Specificity (CI)** | 0.72 (0.62-0.82) | 0.65 (0.55-0.78) | 0.78 (0.70-0.85) | 0.88 (0.85-0.93) | 0.86 (0.80-0.93) | 0.93 (0.89-0.96) |
| **PPV (CI)** | 0.65 (0.56-0.75) | 0.60 (0.52-0.69) | 0.71 (0.61-0.79) | 0.09 (0.02-0.17) | 0.07 (0.02-0.15) | 0.13 (0.03-0.25) |
| **NPV (CI)** | 0.75 (0.68-0.81) | 0.73 (0.66-0.8) | 0.76 (0.70-0.82) | 0.99 (0.99-1.0) | 0.99 (0.99-1.0) | 1.00 (0.99-1.0) |
| **F1-score (CI)** | 0.66 (0.59-0.73) | 0.64 (0.57-0.7) | 0.69 (0.62-0.76) | 0.15 (0.04-0.28) | 0.13 (0.03-0.25) | 0.21 (0.06-0.38) |
| **Accuracy (CI)** | 0.70 (0.65-0.76) | 0.67 (0.61-0.72) | 0.74 (0.68-0.79) | 0.92 (0.88-0.96) | 0.90 (0.86-0.94) | 0.95 (0.92-0.98) |

Note: AUROC: area under the receiver operating characteristic curve; PPV: positive predictive value; NPV: negative predictive value; CI: confidence interval.

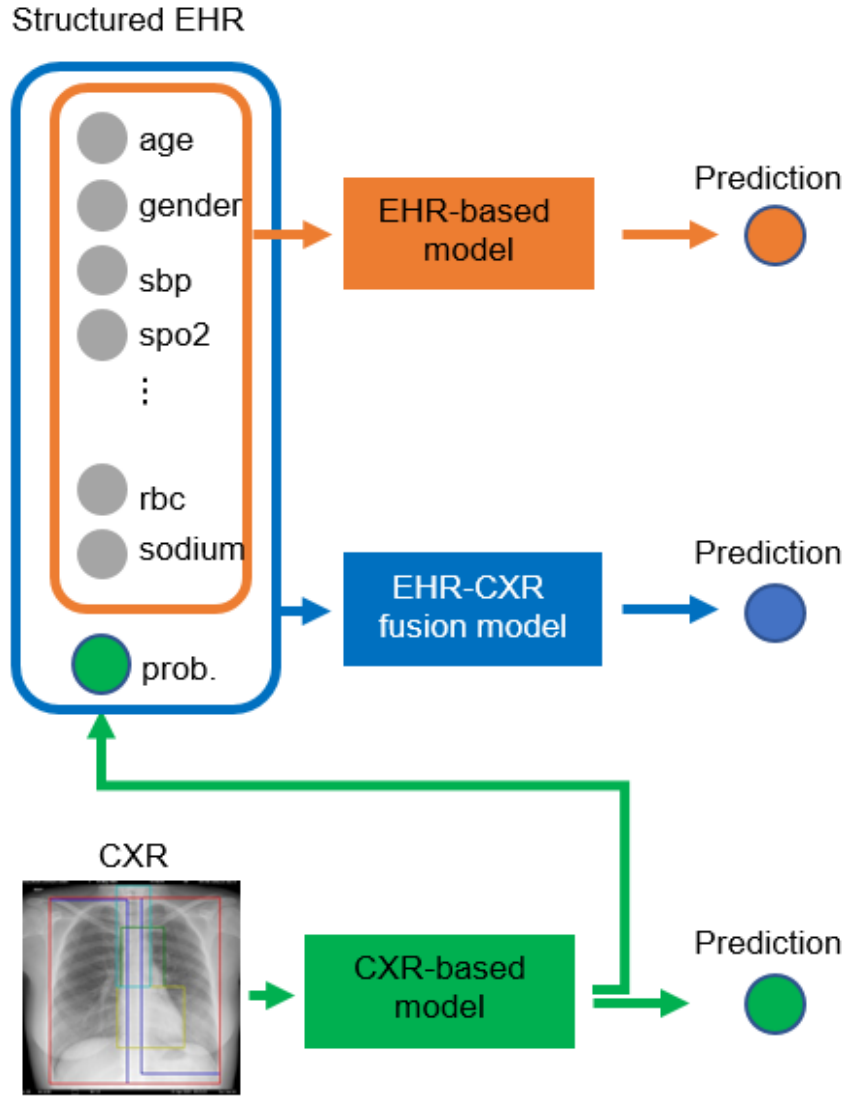

**Figure 1.** The proposed multi-modal models for mortality prediction. EHR data were used to train the EHR-based model and CXR images with augmentation were used to train the CXR-based model. The probability computed from CXR-based model along with EHR data were used for EHR-CXR fusion model.

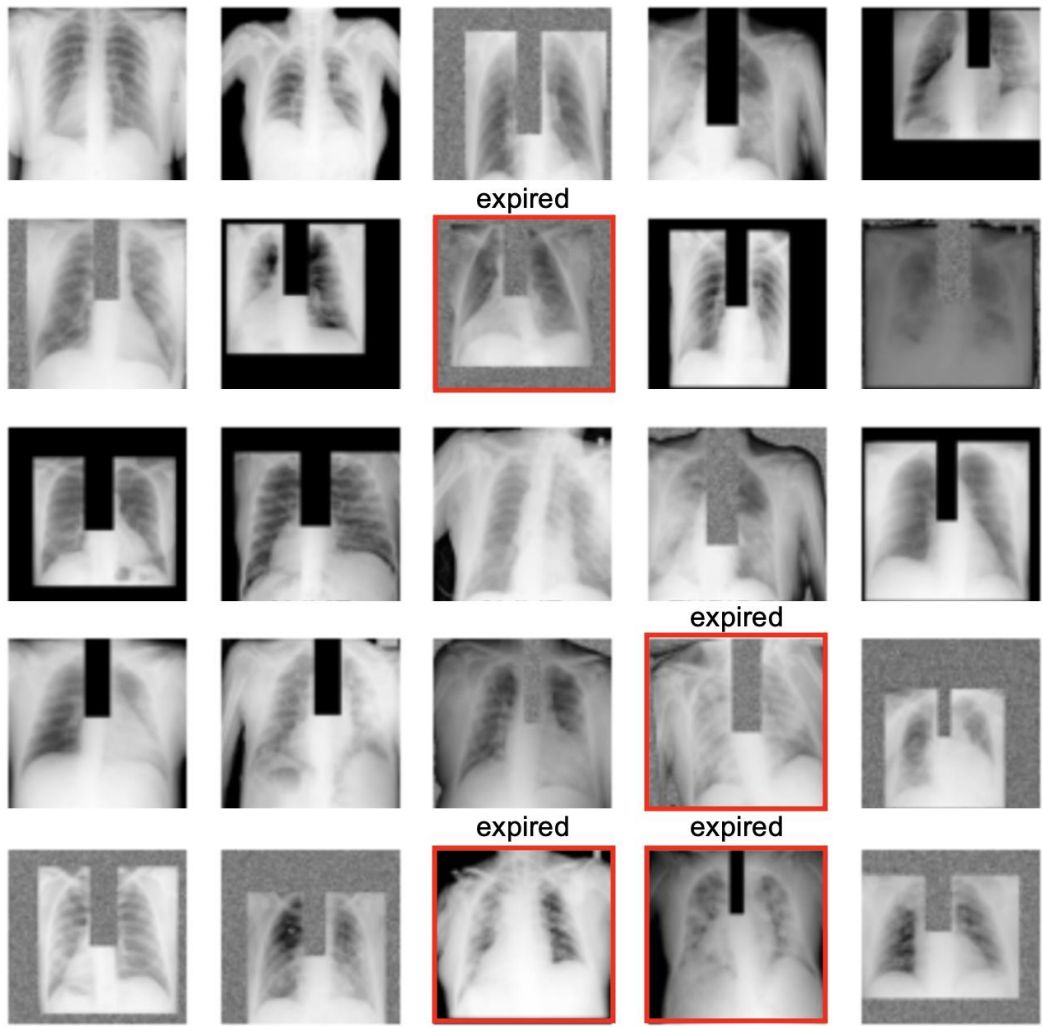

**Figure 2.** A random sample of images shown to teach the model where at least 1-2 positive mortality (expired) cases are shown to the model in each batch.

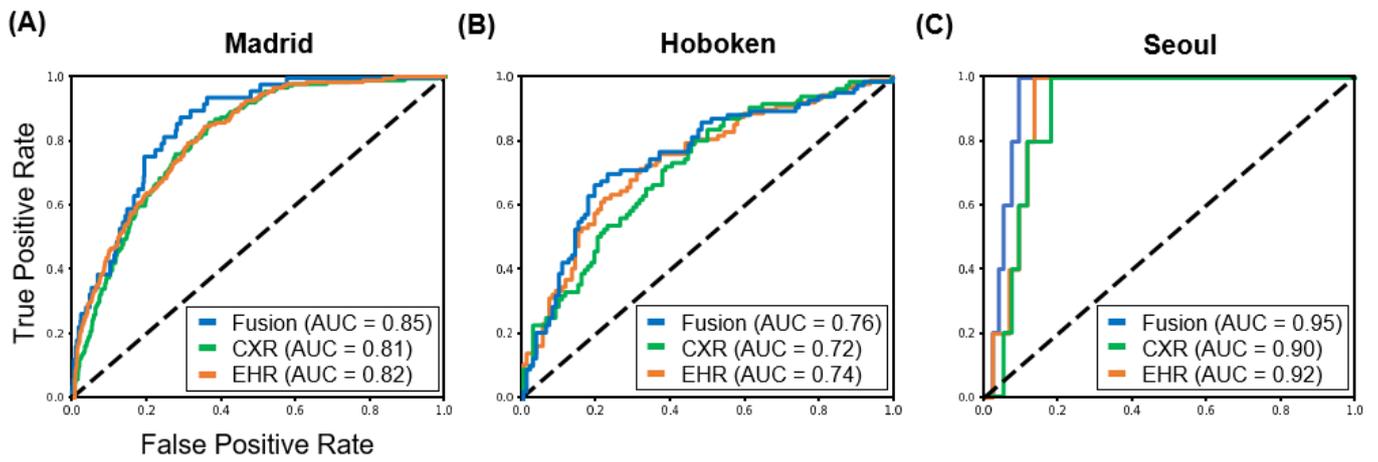

**Figure 3.** Model performance using EHR-based model, CXR-based model, and fusion model (EHR+CXR). (A) Internal validation on Madrid dataset. (B) External testing on Hoboken dataset and (C) External testing on Seoul dataset.

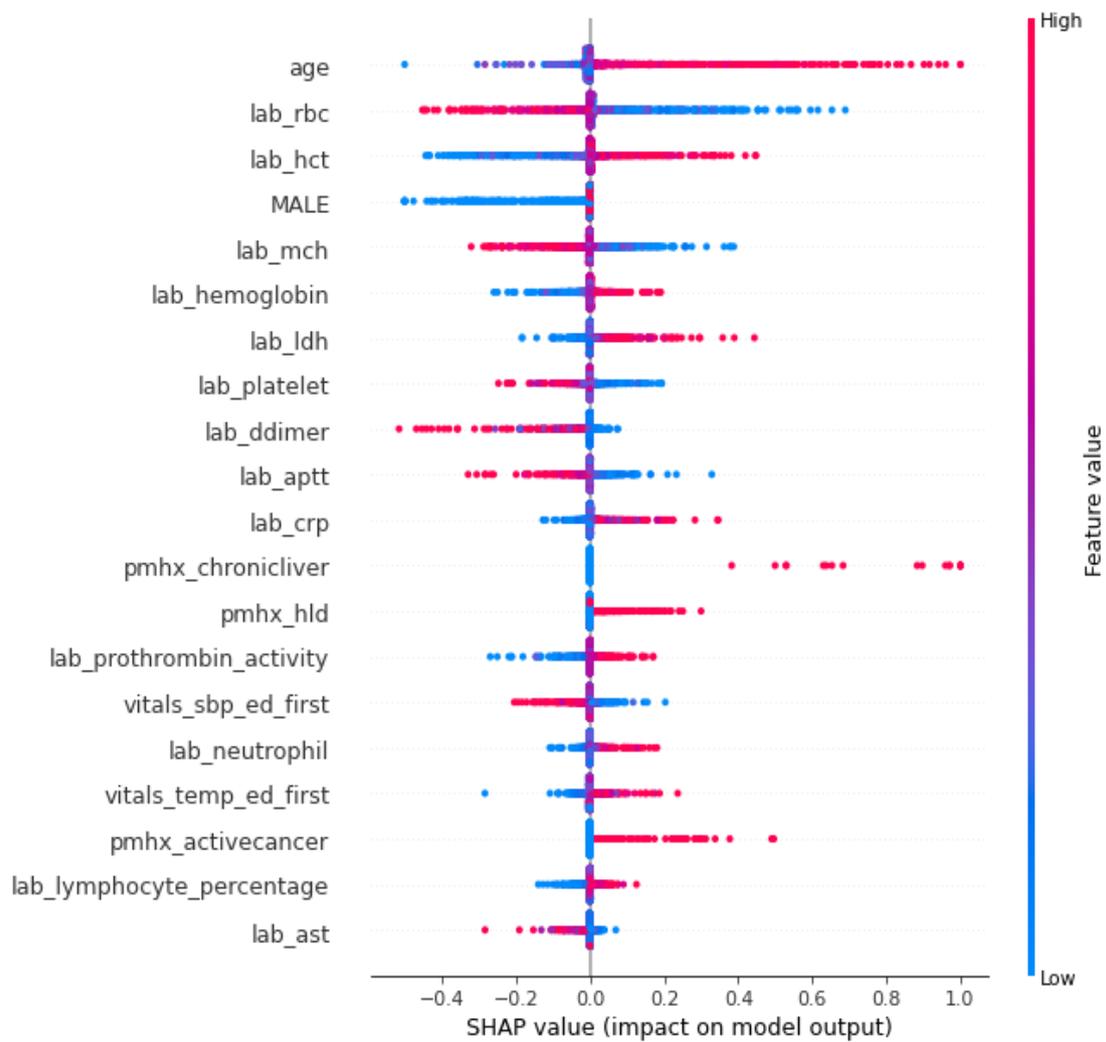

**Figure 4.** Feature importance of the EHR-based model revealed by a SHAP plot. Features on the y-axis are ranked by their mean absolute SHAP values and each point represents a patient.

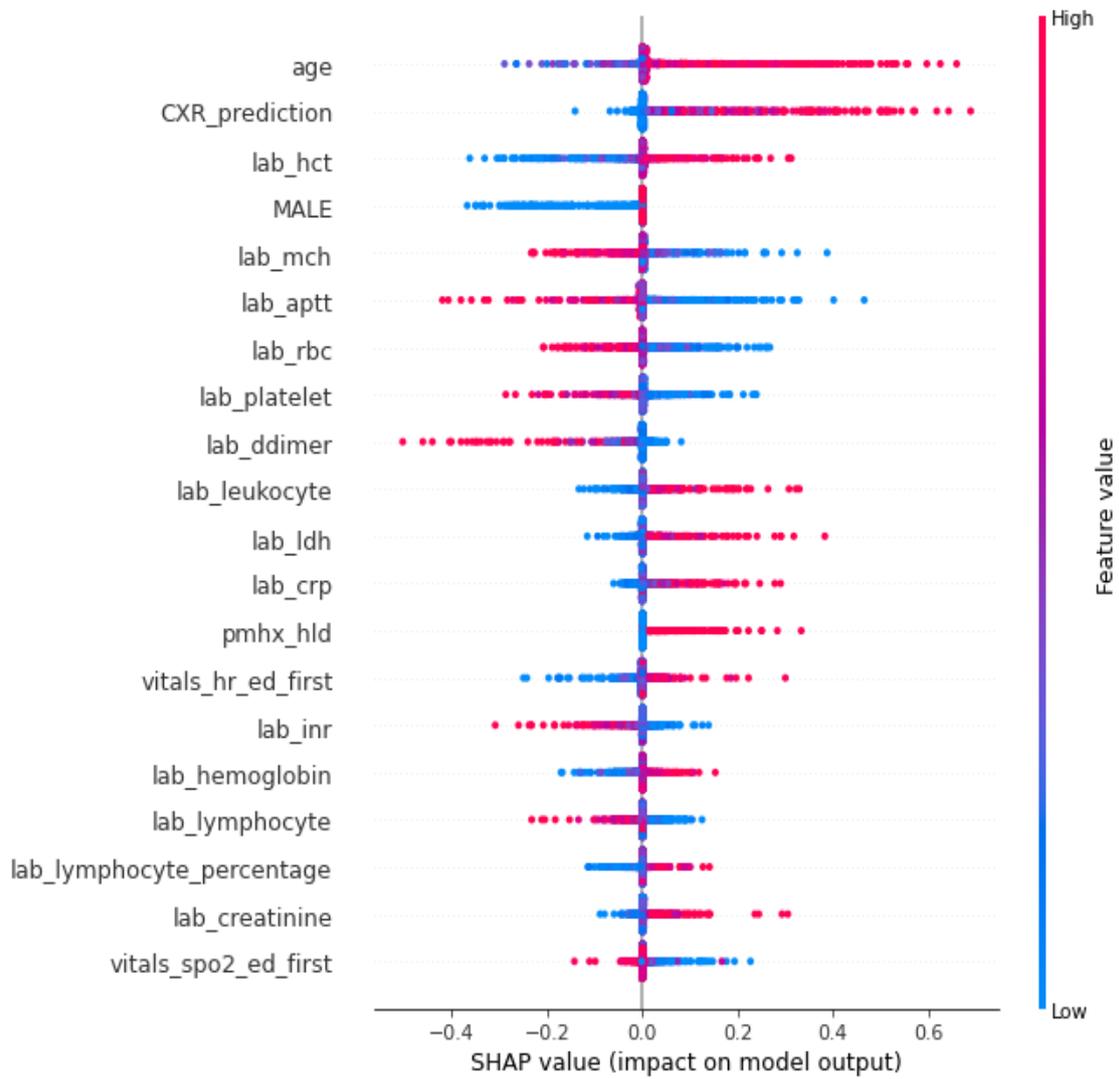

**Figure 5.** Feature importance of the fusion model revealed by a SHAP plot. Features on the y-axis are ranked by their mean absolute SHAP values and each point represents a patient.

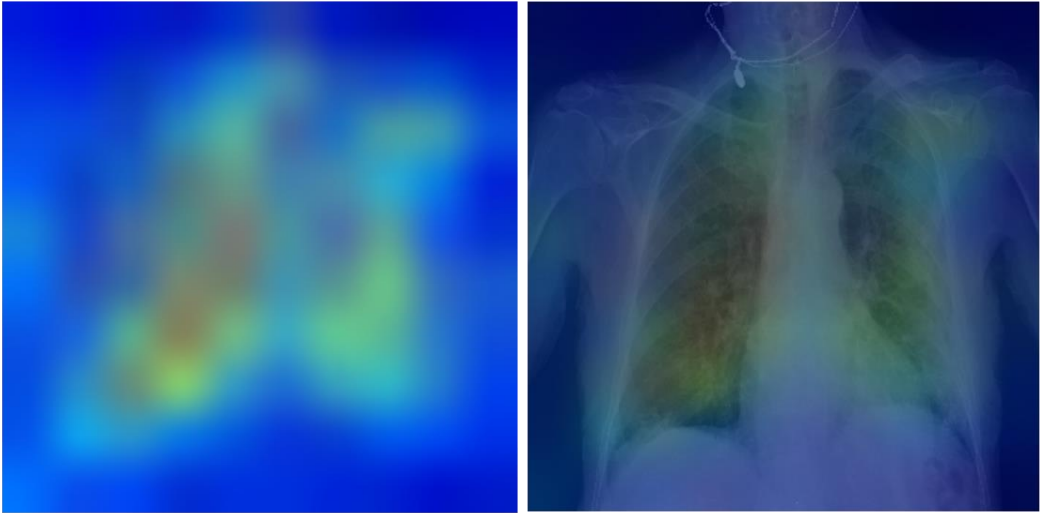

**Figure 6.** Explainability: heatmaps using Grad-CAM algorithm shows that the model primarily uses imaging features from the lungs and mediastinum region for mortality prediction. The image was produced by averaging the heatmaps from the expired patients with prediction probability larger than 0·6 and overlaying it on an actual CXR.

# Supplementary Material

## Part 1 - CLAIM: Checklist for Artificial Intelligence in Medical Imaging[1]

| Section / Topic | No. | Item | Done |
|---|---|---|---|
| TITLE / ABSTRACT | | | |
| | 1 | Identification as a study of AI methodology, specifying the category of technology used (e.g., deep learning) | yes |
| | 2 | Structured summary of study design, methods, results, and conclusions | n/a |
| INTRODUCTION | | | |
| | 3 | Scientific and clinical background, including the intended use and clinical role of the AI approach | yes |
| | 4 | Study objectives and hypotheses | yes |
| METHODS | | | |
| *Study Design* | 5 | Prospective or retrospective study | yes |
| | 6 | Study goal, such as model creation, exploratory study, feasibility study, non-inferiority trial | yes |
| *Data* | 7 | Data sources | yes |
| | 8 | Eligibility criteria: how, where, and when potentially eligible participants or studies were identified (e.g., symptoms, results from previous tests, inclusion in registry, patient-care setting, location, dates) | yes |
| | 9 | Data pre-processing steps | yes |
| | 10 | Selection of data subsets, if applicable | yes |
| | 11 | Definitions of data elements, with references to Common Data Elements | yes |
| | 12 | De-identification methods | yes |
| | 13 | How missing data were handled | yes |
| *Ground Truth* | 14 | Definition of ground truth reference standard, in sufficient detail to allow replication | yes |
| | 15 | Rationale for choosing the reference standard (if alternatives exist) | yes |
| | 16 | Source of ground-truth annotations; qualifications and preparation of annotators | n/a |
| | 17 | Annotation tools | yes |
| | 18 | Measurement of inter- and intra-rater variability; methods to mitigate variability and/or resolve discrepancies | n/a |
| *Data Partitions* | 19 | Intended sample size and how it was determined | n/a |
| | 20 | How data were assigned to partitions; specify proportions | yes |
| | 21 | Level at which partitions are disjoint (e.g., image, study, patient, institution) | yes |
| *Model* | 22 | Detailed description of model, including inputs, outputs, all intermediate layers and connections | yes |
| | 23 | Software libraries, frameworks, and packages | yes |
| | 24 | Initialization of model parameters (e.g., randomization, transfer learning) | yes |
| *Training* | 25 | Details of training approach, including data augmentation, hyperparameters, number of models trained | yes |
| | 26 | Method of selecting the final model | yes |
| | 27 | Ensembling techniques, if applicable | yes |
| *Evaluation* | 28 | Metrics of model performance | yes |
| | 29 | Statistical measures of significance and uncertainty (e.g., confidence intervals) | yes |
| | 30 | Robustness or sensitivity analysis | yes |
| | 31 | Methods for explainability or interpretability (e.g., saliency maps), and how they were validated | yes |
| | 32 | Validation or testing on external data | yes |
| RESULTS | | | |
| *Data* | 33 | Flow of participants or cases, using a diagram to indicate inclusion and exclusion | yes |
| | 34 | Demographic and clinical characteristics of cases in each partition | yes |
| *Model performance* | 35 | Performance metrics for optimal model(s) on all data partitions | yes |
| | 36 | Estimates of diagnostic accuracy and their precision (such as 95% confidence intervals) | yes |
| | 37 | Failure analysis of incorrectly classified cases | no |
| DISCUSSION | | | |
| | 38 | Study limitations, including potential bias, statistical uncertainty, and generalizability | yes |
| | 39 | Implications for practice, including the intended use and/or clinical role | yes |
| OTHER INFORMATION | | | |
| | 40 | Registration number and name of registry | n/a |
| | 41 | Where the full study protocol can be accessed | n/a |
| | 42 | Sources of funding and other support; role of funders | yes |

**Part 2 - Detailed patient characteristics tables**

For statistical analysis, hypothesis test functions used by default are chi-square test for categorical variables (with continuity correction) and one-way test for continuous variables (with equal variance assumption, i.e., regular ANOVA). Two-group ANOVA is equivalent to the t-test. Statistical analyses were performed in Python version 3.7.10. All tests were two-sided, with a 0.05 level of significance.

**Supplementary Table 1. Patient Characteristics for the Madrid dataset**

|  |  | Missing | Overall | Alive | Expired | P-Value |
|---|---|---|---|---|---|---|
| n |  |  | 1628 | 1439 | 189 |  |
| Age, mean (SD) |  | 0 | 67.3 (15.8) | 65.7 (15.7) | 79.6 (10.0) | <0.001 |
| Sex (%) | FEMALE | 0 | 648 (39.8) | 594 (41.3) | 54 (28.6) | <0.001 |
| Temperature (SD) |  | 388 | 36.8 (0.8) | 36.8 (0.8) | 36.8 (0.9) | 0.698 |
| Systolic BP (SD) |  | 604 | 131.0 (21.4) | 131.1 (20.8) | 130.8 (25.1) | 0.916 |
| Diastolic BP (SD) |  | 600 | 75.9 (34.2) | 76.3 (36.2) | 73.2 (14.1) | 0.072 |
| Heart Rate (SD) |  | 368 | 89.7 (16.4) | 89.8 (16.4) | 89.0 (16.4) | 0.558 |
| O2 Saturation (SD) |  | 354 | 92.7 (6.6) | 93.3 (5.8) | 88.3 (9.8) | <0.001 |
| Mortality Outcome, n (%) | FALSE | 0 | 1439 (88.4) | 1439 (100.0) |  | <0.001 |
|  | TRUE |  | 189 (11.6) |  | 189 (100.0) |  |
| Diabetes, n (%) | 1.0 | 9 | 1344 (83.0) | 1202 (83.9) | 142 (75.9) | 0.008 |
| Hyperlipidemia, n (%) | 1.0 | 9 | 439 (27.1) | 373 (26.0) | 66 (35.3) | 0.010 |
| Hypertension, n (%) | 1.0 | 9 | 107 (6.6) | 88 (6.1) | 19 (10.2) | 0.055 |
| Ischemic Heart Disease, n (%) | 1.0 | 9 | 111 (6.9) | 86 (6.0) | 25 (13.4) | <0.001 |
| Chronic Kidney Disease, n (%) | 1.0 | 9 | 95 (5.9) | 73 (5.1) | 22 (11.8) | <0.001 |
| Chronic Obstructive Pulmonary Disease, n (%) | 1.0 | 9 | 83 (5.1) | 68 (4.7) | 15 (8.0) | 0.083 |
| Bronchial Asthma, n (%) | 1.0 | 9 | 83 (5.1) | 75 (5.2) | 8 (4.3) | 0.702 |
| Active Cancer, n (%) | 1.0 | 9 | 80 (4.9) | 57 (4.0) | 23 (12.3) | <0.001 |
| Chronic Liver Disease, n (%) | 1.0 | 9 | 18 (1.1) | 9 (0.6) | 9 (4.8) | <0.001 |
| Previous Stroke, n (%) | 1.0 | 9 | 51 (3.2) | 37 (2.6) | 14 (7.5) | <0.001 |
| Congestive Heart Failure, n (%) | 1.0 | 9 | 75 (4.6) | 62 (4.3) | 13 (7.0) | 0.156 |
| Dementia, n (%) | 1.0 | 9 | 52 (3.2) | 40 (2.8) | 12 (6.4) | 0.015 |
| LDH, mean (SD) |  | 388 | 594.5 (330.8) | 568.1 (257.2) | 805.9 (638.2) | <0.001 |
| MCV, mean (SD) |  | 318 | 88.0 (5.5) | 87.8 (5.3) | 89.4 (6.5) | 0.003 |
| Neutrophil %, mean (SD) |  | 319 | 72.8 (12.1) | 72.1 (11.9) | 78.2 (12.3) | <0.001 |
| aPTT, mean (SD) |  | 769 | 33.0 (6.8) | 32.9 (6.6) | 33.7 (8.1) | 0.359 |
| D-dimer, mean (SD) |  | 600 | 1789.2 (4490.1) | 1694.7 (4448.3) | 2620.6 (4784.2) | 0.061 |
| INR, mean (SD) |  | 641 | 1.4 (1.2) | 1.4 (1.1) | 1.5 (1.3) | 0.392 |
| Glucose, mean (SD) |  | 389 | 125.7 (43.3) | 123.0 (38.5) | 146.1 (66.1) | <0.001 |
| Blood Urea Nitrogen, mean (SD) |  | 367 | 42.4 (32.8) | 39.0 (25.8) | 67.7 (58.7) | <0.001 |
| Lymphocyte %, mean (SD) |  | 319 | 18.4 (9.7) | 18.9 (9.6) | 14.1 (9.4) | <0.001 |
| MCH, mean (SD) |  | 318 | 29.6 (2.0) | 29.6 (2.0) | 29.9 (2.3) | 0.124 |
| AST, mean (SD) |  | 461 | 43.7 (33.2) | 42.7 (33.2) | 51.0 (32.1) | 0.005 |
| RDW, mean (SD) |  | 327 | 12.8 (2.1) | 12.7 (2.0) | 13.5 (2.3) | <0.001 |
| RBC count, mean (SD) |  | 318 | 4.7 (0.6) | 4.7 (0.6) | 4.6 (0.7) | 0.073 |
| Neutrophil, mean (SD) |  | 318 | 5.4 (3.3) | 5.2 (2.9) | 7.0 (5.0) | <0.001 |
| Hemoglobin, mean (SD) |  | 318 | 13.8 (1.8) | 13.8 (1.8) | 13.6 (2.2) | 0.352 |
| CRP, mean (SD) |  | 329 | 92.8 (89.5) | 87.3 (85.9) | 136.2 (104.5) | <0.001 |
| ALT, mean (SD) |  | 480 | 37.7 (35.2) | 38.1 (35.8) | 34.6 (30.8) | 0.220 |
| Creatinine, mean (SD) |  | 337 | 1.0 (0.6) | 0.9 (0.6) | 1.3 (0.8) | <0.001 |
| Mean Platelet Volume, mean (SD) |  | 332 | 10.3 (1.0) | 10.3 (1.0) | 10.6 (1.0) | <0.001 |
| Platelet, mean (SD) |  | 318 | 217.1 (94.7) | 219.8 (92.2) | 197.0 (110.0) | 0.015 |
| Prothrombin Activity, mean (SD) |  | 641 | 74.5 (16.8) | 74.7 (16.5) | 72.8 (18.9) | 0.296 |
| Leukocyte, mean (SD) |  | 318 | 7.1 (3.6) | 6.9 (3.4) | 8.4 (4.7) | <0.001 |
| Serum Sodium, mean (SD) |  | 346 | 136.9 (4.5) | 136.8 (4.0) | 137.5 (7.3) | 0.292 |
| Lymphocyte, mean (SD) |  | 319 | 1.2 (1.3) | 1.2 (1.3) | 0.9 (0.5) | <0.001 |
| Hematocrit, mean (SD) |  | 318 | 40.9 (5.0) | 40.9 (4.8) | 40.8 (6.0) | 0.831 |
| Serum Potassium, mean (SD) |  | 360 | 4.2 (0.5) | 4.2 (0.5) | 4.3 (0.7) | 0.323 |

LDH: lactate dehydrogenase, MCV: mean corpuscular volume, aPTT: partial thromboplastin time, INR: international normalized ratio, MCH: mean corpuscular hemoglobin , AST: aspartate transaminase, RDW: red cell distribution width, RBC: red blood cell, CRP: C-reactive protein, ALT: alanine transaminase

**Supplementary Table 2. Patient characteristics in Hoboken and Seoul Cohorts**

| | | Hoboken University Medical Center | | | | Seoul National University Hospital | | | |
|---|---|---|---|---|---|---|---|---|---|
| | | Overall | Alive | Expired | p-value | Overall | Alive | Expired | p-value |
| n | | 201 | 114 | 87 | | 315 | 310 | 5 | |
| Age, mean (SD) | | 65.0 (16.9) | 61.9 (16.5) | 69.1 (16.5) | 0.003 | 46.0 (19.6) | 45.7 (19.6) | 64.0 (15.2) | 0.053 |
| Sex (%) | F | 83 (41.3) | 55 (48.2) | 28 (32.2) | 0.032 | 150 (47.6) | 150 (48.4) | | 0.062 |
| Temperature (SD) | | 38.0 (0.9) | 38.0 (0.9) | 38.0 (0.9) | 0.853 | 36.7 (0.7) | 36.7 (0.7) | 37.0 (0.6) | 0.320 |
| Systolic BP (SD) | | 129.6 (24.0) | 131.9 (23.1) | 126.5 (25.0) | 0.120 | 122.6 (18.5) | 122.1 (18.1) | 151.6 (22.0) | 0.039 |
| Diastolic BP (SD) | | 73.9 (14.1) | 75.7 (13.9) | 71.5 (13.9) | 0.033 | 86.3 (26.0) | 85.9 (26.9) | 91.4 (9.0) | 0.323 |
| Heart Rate (SD) | | 105.0 (19.7) | 103.8 (16.9) | 106.6 (22.9) | 0.345 | 96.7 (2.5) | 96.8 (2.4) | 92.8 (5.4) | 0.176 |
| O2 Saturation (SD) | | 86.7 (12.5) | 91.1 (8.2) | 81.0 (14.7) | <0.001 | 96.7 (2.5) | 96.8 (2.4) | 92.8 (5.4) | 0.176 |
| Mortality Outcome, n (%) | 0 | 114 (56.7) | 114 (100.0) | | <0.001 | 310 (98.4) | 310 (100.0) | | <0.001 |
| | 1 | 87 (43.3) | | 87 (100.0) | | 5 (1.6) | | 5 (100.0) | |
| Diabetes, n (%) | 1 | 76 (37.8) | 45 (39.5) | 31 (35.6) | 0.682 | 36 (11.5) | 34 (11.0) | 2 (40.0) | 0.103 |
| Hyperlipidemia, n (%) | 1 | 67 (33.3) | 37 (32.5) | 30 (34.5) | 0.880 | 20 (6.4) | 19 (6.2) | 1 (20.0) | 0.283 |
| Hypertension, n (%) | 1 | 110 (54.7) | 62 (54.4) | 48 (55.2) | 0.974 | 47 (15.0) | 43 (14.0) | 4 (80.0) | 0.002 |
| Ischemic Heart Disease, n (%) | 1 | 32 (15.9) | 19 (16.7) | 13 (14.9) | 0.891 | 8 (2.6) | 7 (2.3) | 1 (20.0) | 0.122 |
| Chronic Kidney Disease, n (%) | 1 | 26 (12.9) | 8 (7.0) | 18 (20.7) | 0.008 | 5 (1.6) | 4 (1.3) | 1 (20.0) | 0.078 |
| COPD, n (%) | 1 | 20 (10.0) | 11 (9.6) | 9 (10.3) | 0.941 | 3 (1.0) | 2 (0.6) | 1 (20.0) | 0.047 |
| Bronchial Asthma, n (%) | 1 | 25 (12.4) | 19 (16.7) | 6 (6.9) | 0.062 | 2 (0.6) | 2 (0.6) | | 1.000 |
| Active Cancer, n (%) | 1 | 10 (5.0) | 7 (6.1) | 3 (3.4) | 0.519 | | | | |
| CLD, n (%) | 1 | 1 (0.5) | 1 (0.9) | | 1.000 | 3 (1.0) | 3 (1.0) | | 1.000 |
| Previous Stroke, n (%) | 1 | 7 (3.5) | 3 (2.6) | 4 (4.6) | 0.469 | 8 (2.6) | 8 (2.6) | | 1.000 |
| Congestive Heart Failure, n (%) | 1 | 32 (15.9) | 19 (16.7) | 13 (14.9) | 0.891 | 6 (1.9) | 5 (1.6) | 1 (20.0) | 0.093 |
| Dementia, n (%) | 1 | 28 (13.9) | 9 (7.9) | 19 (21.8) | 0.009 | 3 (1.0) | 3 (1.0) | | 1.000 |
| LDH, mean (SD) | | 1324.4 (1756.1) | 957.1 (372.6) | 1768.9 (2515.5) | 0.005 | 0.0 (0.0) | 0.0 (0.0) | 0.0 (0.0) | nan |
| MCV, mean (SD) | | 87.5 (5.6) | 87.3 (5.5) | 87.9 (5.7) | 0.417 | 0.0 (0.0) | 0.0 (0.0) | 0.0 (0.0) | nan |
| Neutrophil %, mean (SD) | | 82.1 (30.9) | 75.6 (15.3) | 89.9 (41.6) | 0.003 | 0.0 (0.0) | 0.0 (0.0) | 0.0 (0.0) | nan |
| aPTT, mean (SD) | | 33.3 (5.7) | 32.6 (4.4) | 34.0 (6.8) | 0.110 | 0.0 (0.0) | 0.0 (0.0) | 0.0 (0.0) | nan |
| D-dimer, mean (SD) | | 0.0 (0.0) | 0.0 (0.0) | 0.0 (0.0) | | 0.0 (0.0) | 0.0 (0.0) | 0.0 (0.0) | nan |
| INR, mean (SD) | | 1.2 (0.2) | 1.2 (0.2) | 1.2 (0.2) | 0.035 | 0.0 (0.0) | 0.0 (0.0) | 0.0 (0.0) | nan |
| Glucose, mean (SD) | | 165.9 (90.1) | 153.6 (86.5) | 180.7 (92.5) | 0.039 | 0.0 (0.0) | 0.0 (0.0) | 0.0 (0.0) | nan |
| Blood Urea Nitrogen, mean (SD) | | 31.8 (29.7) | 23.3 (21.9) | 42.2 (34.4) | <0.001 | 0.0 (0.0) | 0.0 (0.0) | 0.0 (0.0) | nan |
| Lymphocyte %, mean (SD) | | 12.5 (10.8) | 14.6 (10.7) | 10.0 (10.3) | 0.003 | 0.0 (0.0) | 0.0 (0.0) | 0.0 (0.0) | nan |
| MCH, mean (SD) | | 28.8 (2.2) | 28.8 (2.2) | 28.9 (2.3) | 0.826 | 0.0 (0.0) | 0.0 (0.0) | 0.0 (0.0) | nan |
| AST, mean (SD) | | 102.1 (252.0) | 74.3 (70.6) | 135.7 (364.6) | 0.125 | 0.0 (0.0) | 0.0 (0.0) | 0.0 (0.0) | nan |
| RDW, mean (SD) | | 14.8 (2.0) | 14.5 (1.9) | 15.1 (2.0) | 0.045 | 0.0 (0.0) | 0.0 (0.0) | 0.0 (0.0) | nan |
| RBC count, mean (SD) | | 4.6 (0.7) | 4.6 (0.6) | 4.6 (0.7) | 0.674 | 0.0 (0.0) | 0.0 (0.0) | 0.0 (0.0) | nan |
| Neutrophil, mean (SD) | | 8.4 (8.7) | 6.7 (4.2) | 10.5 (11.8) | 0.005 | 0.0 (0.0) | 0.0 (0.0) | 0.0 (0.0) | nan |
| Hemoglobin, mean (SD) | | 14.0 (10.4) | 13.4 (1.9) | 14.8 (15.3) | 0.378 | 0.0 (0.0) | 0.0 (0.0) | 0.0 (0.0) | nan |
| CRP, mean (SD) | | 166.4 (120.3) | 114.2 (90.9) | 231.5 (121.1) | <0.001 | 0.0 (0.0) | 0.0 (0.0) | 0.0 (0.0) | nan |
| ALT, mean (SD) | | 66.1 (149.1) | 52.2 (62.9) | 82.6 (208.9) | 0.193 | 0.0 (0.0) | 0.0 (0.0) | 0.0 (0.0) | nan |
| Creatinine, mean (SD) | | 2.0 (3.7) | 1.3 (2.2) | 2.7 (4.9) | 0.019 | 0.0 (0.0) | 0.0 (0.0) | 0.0 (0.0) | nan |
| Mean Platelet Volume, mean (SD) | | 8.5 (1.0) | 8.4 (0.9) | 8.5 (1.0) | 0.499 | 0.0 (0.0) | 0.0 (0.0) | 0.0 (0.0) | nan |
| Platelet, mean (SD) | | 242.3 (96.6) | 240.6 (89.4) | 244.4 (105.1) | 0.793 | 0.0 (0.0) | 0.0 (0.0) | 0.0 (0.0) | nan |
| Prothrombin Activity, mean (SD) | | 13.8 (2.2) | 13.5 (2.1) | 14.2 (2.4) | 0.093 | 0.0 (0.0) | 0.0 (0.0) | 0.0 (0.0) | nan |
| Leukocyte, mean (SD) | | 10.0 (8.6) | 9.0 (8.5) | 11.1 (8.8) | 0.095 | 0.0 (0.0) | 0.0 (0.0) | 0.0 (0.0) | nan |
| Serum Sodium, mean (SD) | | 136.7 (7.0) | 135.3 (4.6) | 138.5 (8.8) | 0.003 | 0.0 (0.0) | 0.0 (0.0) | 0.0 (0.0) | nan |
| Lymphocyte, mean (SD) | | 1.3 (5.0) | 1.6 (6.7) | 1.0 (1.3) | 0.338 | 0.0 (0.0) | 0.0 (0.0) | 0.0 (0.0) | nan |
| Hematocrit, mean (SD) | | 40.4 (5.7) | 40.5 (5.5) | 40.3 (6.0) | 0.870 | 0.0 (0.0) | 0.0 (0.0) | 0.0 (0.0) | nan |
| Serum Potassium, mean (SD) | | 4.7 (6.6) | 5.0 (8.9) | 4.3 (0.6) | 0.364 | 0.0 (0.0) | 0.0 (0.0) | 0.0 (0.0) | nan |

COPD: Chronic Obstructive Pulmonary Disease, CLD: Chronic Liver Disease, LDH: lactate dehydrogenase, MCV: mean corpuscular volume, aPTT: partial thromboplastin time, INR: international normalized ratio, MCH: mean corpuscular hemoglobin , AST: aspartate transaminase, RDW: red cell distribution width, RBC: red blood cell, CRP: C-reactive protein, ALT: alanine transaminase

**Part 3 - Case Definitions and Ethical Statements**

*Madrid* - From all patients with COVID-19 (N = 2547) admitted at Madrid, the vast majority of patients have been diagnosed by positive PCR. However, during the months of March-April 2020, when there was no PCR test, the diagnosis was made by clinical and/or radiological signs from an CXR and symptoms compatible with bilateral pneumonia.

*Hoboken* - Data of all patients with COVID-19 (N = 242) admitted at the Hoboken University Medical Center until April 11, 2020, were retrospectively collected on April 21, 2020. COVID-19 was confirmed in all patients using quantitative real-time reverse transcription polymerase chain reaction for SARS-CoV-2 RNA. Data for patients who did not meet the primary outcome were excluded on the 30th day of admission.

*Seoul* - Data of all patients with COVID-19 (N = 336) admitted at Seoul are patients diagnosed with COVID-19 (PCR confirmed) from Seoul clinical data warehouse (CDW) and admitted to the intensive care unit at Seoul.

Separate IRBs and data use agreements were independently obtained from the data controller and the ethics board of the source institutions for the three different datasets to conduct this study. Data access to different datasets by researchers in this study is given via an as needed basis after the researchers and their institutions signed the relevant data use agreement. The purpose of the study is non-commercial and the ethical statements for the following datasets are the following:

- *Madrid* - CEIm Ref No. 20.05.1627-GHM Title of the Protocol: Clinical course and outcomes of severe and critical COVID-19 patients on interleukin-6 inhibitors: a retrospective cohort study; protocol identification: Covid-IL6; IRB Sponsor: Fundación de Investigación HM Hospitales.

- *Hoboken* - This patient population was previously reported by Yao et al and the study protocol was approved and was granted a waiver of informed consent by the hospital board on April 15, 2020.[2] Data extraction, collection, and analyses and external model validation on this dataset were performed by two trained physicians from HUMC (JAP and JSY).

- *Seoul* - IRB No. H-2007-065-1140 from Seoul National University Hospital, Republic of Korea.

**Part 4 – Structured EHR data preprocessing**

*Deidentification -* The following identifiers of the individual or of relatives, employers, or household members of the individual, were removed for de-identifying the dataset: Names; all geographic subdivisions smaller than a state, including street address, city, county, precinct, ZIP code, and their equivalent odes, except for the initial three digits of the ZIP code if, according to the current publicly available data from the Bureau of the us, all elements of dates (except year) for dates that are directly related to an individual, including birth date, admission date, discharge date, death date, and all ages over 89 and all elements of dates (including year) indicative of such age, except that such ages elements may be aggregated into a single category of age 90 or older; telephone numbers; vehicle identifiers and serial numbers, including license plate numbers; fax numbers; device identifiers and serial numbers; email addresses; web Universal Resource Locators (URLs); social security numbers; internet Protocol (IP) addresses; medical record numbers; health plan beneficiary numbers; account numbers; any other unique identifying number, characteristic, or code.[3]

*Outliers -* We cleaned up systolic blood pressure (systolic BP), heart rate, SPO2 and temperature to be within valid ranges (code available). The valid ranges used for temperature, SPO2, heart rate, and systolic BP are 30-45 C, 1-100 %, 20-300 bmp, 20-240 mmHg, respectively.

*Missing values* - We removed labs that had more than 50% missing values in the model development dataset (Madrid). The remaining missing values are imputed following common data science procedures, where missing categorical values were filled by most frequent value imputation and continuous values by median imputation. Variables not available in Hoboken or Seoul but available in Madrid were filled with 0s.

*Data normalization* - The categorical variables were transformed by a one-hot encoding method. The numerical variables were scaled based on percentiles across the whole Madrid dataset with the robust scaler method in python package (scikit learn 0.21).

## Part 5 – Design decisions and Reasons

| Study objectives | |
|---|---|
| *Design decisions* | Optimize for F1-score for all three models for the 30-day mortality prediction task and report all metrics including areas under the receiver operating characteristic curve (AUROC), sensitivity, specificity, positive predictive value (PPV), negative predictive value (NPV), F1-score, and accuracy. |
| *Reasons* | 30-day mortality is chosen as the target outcome in accordance with clinical precedence.[4–6] The F1-score finds an equal balance between PPV (precision) and sensitivity (recall), which gives a better indication of model performance for unbalanced dataset (mortality is relatively rare compared to survival). Reporting all metrics allows assessment of how the models might perform in populations with a different COVID-19 related mortality distribution. With COVID mortality rates varying with time, model drift can be a real concern under different care delivery parameters during surges.[7] |

| Datasets | |
|---|---|
| *Design decisions* | We included only cases with admission CXRs and results of laboratory tests taken within the first 24 hours of hospital admission in Table 1 and 2. Cases were also excluded for missing admission time. Patients aged 16 or under are excluded and only frontal (AP or PA) images are included for the CXR-based and EHR-CXR fusion models. |
| *Reasons* | Our clinical goal is to develop an early assessment algorithm. Admission time is needed to establish the 30-day mortality cut off for this study. Patients under 16 need more privacy protection (very rare and more easily re-identifiable) and their CXR imaging appearance (anatomically) and disease outcome distributions are very different. Not all CXR exam orders include lateral images hence they are not included as an input for the models. |

| Data preprocessing | |
|---|---|
| *Design decisions* | An anatomical bounding box (Bbox) extraction pipeline was used to automatically extract the coordinates for the left lung, right lung, mediastinum, and trachea anatomies from each of the frontal CXR images.[8] The extracted bounding boxes are reviewed and manually corrected as needed by clinicians (JAP, JSY, ECD). We used these anatomical Bboxes to create 4 additional versions for each image for augmentation, where in version 1) trachea Bbox was masked out with 0's, 2) trachea Bbox was replaced with random noise, 3) background and trachea Bboxes were masked out with 0's, and 4) background and trachea boxes were replaced with random noise. The original and augmented images are pre-saved as JPEGs without resizing at this stage. During training, when the hyperparameter for 'augment_bbox' is set to true, a random version of each image (including possibly the non-augmented version) is drawn to teach the model in each epoch (see Figure 2). |
| *Reasons* | As compared to simply post-hoc assessing the explainability of models with Gradient-weighted Class Activation Mapping (Grad-CAM), we tried to force the CXR model to learn features from key CXR anatomies that should be relied on more heavily for prediction during the model training stage as well.[8] Non-augmented image examples are also used in the training so that the model can handle non-augmented CXR images too at inference (i.e. clinical deployment setting). Doing the Bbox augmentation offline not only makes training faster but also more deterministic. We left input size for images as a tuning parameter that is dependent on the pre-trained model teacher. |

| Training and model picking | |
|---|---|
| **All models** | |
| *Design decisions* | The whole of Madrid dataset was randomly divided into four subsets in order to conduct a 4-fold cross-validation training strategy to select for the best model (by F1-score) for each of the three model types. We ensured similar numbers of mortality cases in each split and the same four-way split was used for all experiments. Finally, we trained each of the 3 models on all of Madrid data once we identified the best hyperparameters from the 4-fold cross validation hyperparameter tuning experiments. The final models are then validated on the two external datasets (Hoboken and Seoul). See supplementary materials for the details of model training and picking for EHR, CXR, and fusion models. All models are trained on Google Cloud TPUs via Colab notebooks. Code for both training with paid and free TPUs are available. Software packages used were tensorflow==2.4.1, sklearn-pandas==1.8.0, xgboost==0.90. To ensure repeatability, a random seed of 2020 was used for all experiments. |
| *Reasons* | In this setup, for each experiment, 3 folds are combined and used for training and the fourth fold is used for validation, whereby each individual data subset gets equal opportunities to validate models. Model parameters that perform best on one validation subset might just be 'lucky'. Rotating the validation subset and picking model parameters that perform the best on average across all subsets helps in selecting a model that has hopefully learned more reliable features and may generalize better on external validation sets. Similar number of positive mortality cases (expired patients) in each split makes the validation set more likely to be equally difficult. We had to use the whole Madrid dataset for model development (training and validation) otherwise the number of positive cases (mortality) would be too small for tuning. We used only open sourced python packages so that others can easily re-use and build on our work with no cost barriers. |
| **EHR-based model** | |
| *Design decisions* | Four different types of machine learning algorithms (logistic regression, random forest, gradient boosting, and XGBoost) implemented in scikit learn 0.21 were tried in a tuning setting to select for the best EHR-based model. A randomized grid-search method was used to sample different hyperparameter settings from prespecified ranges for each optimization experiment, as shown in Supplementary Table 3. |
| *Reasons* | The goal of this modeling is not causality analysis but simply to select a model that performs the best for the given dataset and prediction task. We picked the four most common machine learning algorithms suitable for modeling tabular data and tuned their hyperparameters. |
| **CXR-based model** | |
| **Step 1: Online (real-time) image augmentation during training** | |
| *Design decisions* | CXR images were randomly flipped vertically (left-right) and brightness adjusted (0-0.05). Together with the preprocessed anatomical Bbox augmentation, a random set of CXRs used for training the model is illustrated in Figure 2. Both the online and offline augmentations are only used during training and not during internal and external validation of models. |
| *Reasons* | The goal of image augmentation is to automatically increase training sample variety so that the model can learn to discern features that are more generalizable for the downstream prediction task. This step is particularly important if the training dataset is small. The online augmentations (flip and brightness) try to simulate how variations under which CXRs can be taken in real life might alter the image appearance. Only small augmentation ranges are chosen so that the CXR images remain radiologically interpretable. Augmentation is not used during internal and external validation because there is 1) no need to update model weights during evaluation settings and 2) need for comparing models against a consistent benchmark and augmentation introduces randomness. |
| **Step 2: Online CXR feature extraction** | |
| *Design decisions* | Two different previously published pre-trained DenseNet-121 CXR models are tried for feature selection for our downstream mortality prediction task. The Madrid CXR images are resized during training to the input size for each of the pre-trained models (320x320 vs 224x224) to output the imaging features for classification. The last fully connected layer of both models, containing 14 outputs corresponding to the 14 radiologic CheXpert finding labels, was removed.[9] Instead, linearized convolutional features from either the second (-2) or the fourth (-4) to the last layer were used for the mortality prediction classification task. The pre-trained models were partially frozen, with model weights updating after either layer 355, 400 or 420 during training. Choices for which 'teacher' pre-trained model, feature layer to use, and how many model layers to update for the new mortality prediction task are set up as hyperparameters to be tuned in our experiments. |
| *Reasons* | The Madrid dataset is too small to train deep learning networks from scratch. The pre-trained CXR models chosen have already been trained on much larger CXR datasets (MIMIC-CXR) (>200,000 images) to discern features that are useful for diagnosing 14 different CXR lung and heart radiologic findings, which are also clinically relevant for COVID patients.[10] The final few layers in pre-trained convolutional neural networks tend to have best summarized the features useful for downstream (related) classification tasks. Since we only have a small training dataset (Madrid), we decided to only partially update the weights in the later layers in the pre-trained models and leave the choice of how many layers to update as a tunable parameter -- knowing that there is a balance to be 'learned' between updating weights for the new task on the small Madrid training dataset and losing the benefit of pre-learned weights from the pre-trained 'teacher' CXR models. |
| **Step 3: Mortality classification layers** | |
| *Design decisions* | After CXR features are extracted from a pre-trained model, we added a classification block consisting of tunable number of hidden linear layers, followed by a final activation function (choice between ReLU and LeakyReLU), a dropout layer, and a single binary output layer. The output layer represents whether a patient is alive or expired at 30-days. An initial bias to the final out layer was optionally added and tuned along with the choices for activation function (ReLU or LeakyReLU) and the number and sizes of the hidden layers. |
| *Reasons* | The feature size extracted from both pre-trained models is 1024 in length. Additional classification layers were added to learn the new mortality classification task. Since the layer numbers and sizes are arbitrary, we picked a few common sizes to tune. We tried LeakyRelu as an activation function in the classification block because the CXR features extracted from the (-2) and (-4) layers can have many zeros due to the DenseNet-121 architecture. Adding initial bias to the output layer can help with performance for very unbalanced dataset. |
| **Step 4: Optimization settings** | |
| *Design decisions* | Binary cross entropy was used as the loss function and the Adam optimizer was used for parameter optimization. We did not tune for these settings. |
| *Reasons* | Binary cross entropy as the loss is appropriate for the binary mortality classification task. Adam is a fast optimizer, helps with avoiding overfitting and has shown good performance over a range of tasks. |
| **Step 5: Hyperparameter tuning and model selection** | |
| *Design decisions* | Supplementary Table 4 provides a summary of all the hyperparameters we experimented with on the Madrid dataset to select for the final best performing CXR-based COVID-19 30-day prediction model. An experiment is defined by one unique combination of hyperparameters. Due to limited training resources and a large hyperparameter search space (345,600 unique combinations), we had to first rough search and manually narrow down the hyper-parameter search space - e.g. early observation suggests most experiments did better with smaller batch sizes, LeakyReLU activation, and with Bbox augmentation. We then fine-tuned the model on the other more important parameters such as the learning rate. Early stopping was used to end experiments that did not show loss reduction after 2 or 5 epochs. Overall, we performed over 300 experiments. For each experiment, we plotted the train and valid curves for multiple metrics (recall, precision, accuracy, AUC and F1-score) against the number of epochs. We performed a range of manual and automatic model selection by 1) evaluating experiments with F1-scores above 0.25 for all four folds, and 2) manually examining the train-vs-validation learning curves to pick the hyperparameter setting that showed improvement of the model's precision and recall from baseline for both the train and valid data, as well as ensuring that the chosen model did not show evidence of over-fitting. |

| | | |
|---|---|---|
| *Reasons* | The standard practice for hyperparameter tuning is to update model weights on the train dataset and evaluate the updated model on the validation dataset at the end of each epoch, which is when the model has 'seen' all examples in the train set once. Despite using all of the Madrid dataset for training and validation, the number of positive cases in the valid set is still small. Simply picking the best F1-score automatically without inspecting all the learning curves could just end up picking a 'lucky' epoch. | |
| **EHR-CXR fusion model** | | |
| *Design decisions* | We took a late fusion approach that uses the output probability from the CXR model as a feature along with the EHR features for the 30-day mortality classification. With the Madrid train dataset, we again tuned four different machine learning models (logistic regression, random forest, gradient boosting, and XGBoost) in a 4-fold cross-validation setting and the best model along with the best hyper-parameters were selected using randomised grid search via the same methodology as that for training the EHR-based model. | |
| *Reasons* | Late fusion approach is used because it can be implemented with traditional machine learning methods, which can avoid overfitting for smaller datasets. On the other hand, intermediate (Joint) fusion implemented by neural networks requires more data for training (the implementation of the intermediate fusion model can also be found in Supplementary Table 6). In addition, the much larger feature size from imaging modality can easily swamp important clinical signals from the tabular EHR data. From analyzing the fusion model's point of view, late fusion allows interpretation of the overall feature importance from the CXR model's prediction. | |
| **Model evaluation** | | |
| *Design decisions* | We made a clear separation between model developers and final model testers. Development of models includes programming feature selection and model training (JTW, PK, WY). External testing of models (JAP, JSY, MA) requires institutional access for the Hoboken and Seoul data, which were obtained upon request with submission of our study protocol. | |
| *Reasons* | This is the best practice to avoid repeated testing on the final test datasets, which could invalidate the reported results. It is also a common setting in real life model evaluation scenarios. | |
| **Code packaging for testing** | | |
| *Design decisions* | We packaged the inference code for the three different models for testing in an end-to-end Colab Notebook for the model testers to run on their datasets. | |
| *Reasons* | All datasets had been de-identified and are hosted on different HIPPA compliant cloud servers with access granted to different researchers based on institutional affiliation, data access approvals and IRBs. Running via Colab, which have access management protocols, allows the clinical researchers to run the inference code without setting up Python and other required packages on their local machines, which can be a technical barrier. | |

**Supplementary Table 3. Structured EHR-based model's hyperparameter search space**

| | Hyperparameter | Choices | No. of choices |
|---|---|---|---|
| Logistic regression* | Alpha | uniform(0.0001,0.001) | range |
| | Penalty | [l1, l2, 'elasticnet'] | 3 |
| | l1_ratio | uniform(0.01,0.30) | range |
| Random forest | bootstrap | [True, False] | 2 |
| | max_depth | randint(3,12) | 10 |
| | max_features | ['auto', 'sqrt'] | 2 |
| | min_samples_split | randint(2,12) | 11 |
| | min_samples_leaf | randint(2,12) | 11 |
| | n_estimators | randint(200, 1000) | 801 |
| Gradient boosting | Loss | ['deviance','exponential'] | 2 |
| | learning_rate | uniform(0.003, 0.3) | range |
| | n_estimators | randint(200, 1000) | 801 |
| | subsample | uniform(0.1, 1) | range |
| | criterion | ['friedman_mse','mse','mae'] | 3 |
| | min_samples_split | randint(2,12) | 11 |
| | min_samples_leaf' | randint(2,12) | 11 |
| | max_depth | randint(3,12) | 10 |
| | max_features | ['sqrt', 'log2'] | 2 |
| XGBoosting | colsample_bytree | uniform(0.1,1) | range |
| | eta | (0.0001,0.1) | 2 |
| | max_depth | randint(3,12) | 10 |
| | min_child_weight | randint(3,12) | 10 |
| | subsample | uniform(0.1,1) | range |

Note: *: The best model were selected with hyperparameters Alpha = 0.0007, Penalty = l1, and l1 ratio = 0.02 for EHR-based model; Alpha = 0.0005, Penalty = l1, and l1 ratio = 0.255 for fusion model.

**Supplementary Table 4. CXR-based model's hyperparameter search space**

| Hyperparameter | Choices | No. of choices | Best param |
|---|---|---|---|
| Bbox augmentation | [True, False] | 2 | True |
| Model teacher | [224x224 CheXNet model, 320x320 MIMIC model][11] | 2 | 320x320 MIMIC model |
| Frozen layers | [0-355, 0-400, 0-420] | 3 | 0-400 |

| Feature layer | [-2, -4] | 2 | -2 |
|---|---|---|---|
| Hidden layers for classifier | [None, [512], [128, 128], [128, 64], [128]] | 5 | [128, 64] |
| Activation function for classifier | [ReLU, LeakyReLU] | 2 | LeakyReLU |
| Output initial bias | [None, calculated initial bias (np.log([expired/alive])] | 2 | Calculated initial bias |
| Batch size | [16, 32] | 2 | 16 |
| Class weight | [1:10, 3:10, 5:10, 8:10, 1:1, calculated weights of 0.56:4.47] | 6 | 1:1 (i.e., no re-weighting) |
| Drop out | [0.1, 0.25, 0.3, 0.4, 0.5] | 5 | 0.5 |
| Initial learning rate (with a fixed exponential decay rate of 0.96) | [0.001, 0.002, 0.003, 0.005, 0.01, 0.015] | 6 | 0.002 |
| Number of epochs | Chosen by early stopping with patience of 2 or 5 epochs up to maximum of 30 epochs (fixed) | 2 | 19 epochs with patience of 2 |

**Part 6 - Fairness analysis details**

**Supplementary Table 5. Evaluation metrics of fairness analysis between male and female patients.**

| | | | Male | Female |
|---|---|---|---|---|
| **Madrid** | EHR-based | AUROC | 0.83 [0.80-0.86] | 0.77 [0.72-0.82] |
| | | Sensitivity | 0.81 [0.73-0.86] | 0.72 [0.61-0.82] |
| | | Specificity | 0.71 [0.66-0.78] | 0.66 [0.57-0.76] |
| | | PPV | 0.29 [0.25-0.34] | 0.15 [0.11-0.2] |
| | | NPV | 0.96 [0.95-0.97] | 0.97 [0.95-0.98] |
| | | F1-score | 0.43 [0.38-0.47] | 0.24 [0.19-0.3] |
| | | Accuracy | 0.72 [0.68-0.77] | 0.66 [0.58-0.75] |
| | CXR-based | AUROC | 0.79 [0.76-0.82] | 0.81 [0.78-0.83] |
| | | Sensitivity | 0.75 [0.68-0.83] | 0.76 [0.71-0.82] |
| | | Specificity | 0.71 [0.64-0.76] | 0.72 [0.67-0.75] |
| | | PPV | 0.28 [0.23-0.33] | 0.25 [0.21-0.28] |
| | | NPV | 0.95 [0.94-0.96] | 0.96 [0.95-0.97] |
| | | F1-score | 0.40 [0.35-0.45] | 0.37 [0.33-0.41] |
| | | Accuracy | 0.72 [0.66-0.76] | 0.73 [0.68-0.76] |
| | Fusion | AUROC | 0.84 [0.81-0.86] | 0.85 [0.81-0.88] |
| | | Sensitivity | 0.78 [0.72-0.84] | 0.79 [0.71-0.87] |
| | | Specificity | 0.74 [0.68-0.79] | 0.76 [0.69-0.82] |
| | | PPV | 0.31 [0.26-0.36] | 0.21 [0.16-0.27] |
| | | NPV | 0.96 [0.95-0.97] | 0.98 [0.97-0.99] |
| | | F1-score | 0.44 [0.39-0.49] | 0.33 [0.26-0.41] |
| | | Accuracy | 0.75 [0.70-0.78] | 0.76 [0.70-0.82] |
| **Hoboken** | EHR-based | AUROC | 0.74 [0.67-0.82] | 0.73 [0.62-0.83] |
| | | Sensitivity | 0.68 [0.57-0.79] | 0.67 [0.52-0.81] |
| | | Specificity | 0.73 [0.62-0.84] | 0.71 [0.58-0.84] |
| | | PPV | 0.72 [0.62-0.83] | 0.55 [0.40-0.71] |
| | | NPV | 0.70 [0.60-0.79] | 0.81 [0.71-0.9] |
| | | F1-score | 0.70 [0.61-0.78] | 0.60 [0.48-0.71] |
| | | Accuracy | 0.71 [0.64-0.77] | 0.70 [0.60-0.78] |
| | CXR-based | AUROC | 0.77 [0.69-0.84] | 0.66 [0.55-0.76] |
| | | Sensitivity | 0.76 [0.62-0.9] | 0.65 [0.48-0.85] |
| | | Specificity | 0.70 [0.58-0.82] | 0.64 [0.44-0.82] |
| | | PPV | 0.72 [0.62-0.81] | 0.49 [0.35-0.65] |
| | | NPV | 0.74 [0.63-0.88] | 0.78 [0.67-0.88] |
| | | F1-score | 0.74 [0.65-0.81] | 0.55 [0.44-0.66] |
| | | Accuracy | 0.73 [0.66-0.8] | 0.64 [0.54-0.75] |
| | Fusion | AUROC | 0.75 [0.67-0.82] | 0.69 [0.57-0.79] |
| | | Sensitivity | 0.66 [0.53-0.8] | 0.65 [0.50-0.8] |
| | | Specificity | 0.75 [0.58-0.9] | 0.70 [0.56-0.84] |
| | | PPV | 0.73 [0.61-0.87] | 0.53 [0.38-0.69] |
| | | NPV | 0.68 [0.59-0.78] | 0.80 [0.70-0.89] |
| | | F1-score | 0.69 [0.60-0.77] | 0.58 [0.46-0.69] |
| | | Accuracy | 0.70 [0.64-0.77] | 0.68 [0.59-0.77] |

Note: No cases of death amongst females in the Seoul dataset.

**Part 7 - Project code and resources**

The authors provided open access to all their data extraction, filtering, data wrangling, modeling, figures and tables, code, and queries on https://github.com/theonesp/multimodal_mortality_covid . The de-identified version of Madrid COVID Data Saves Lives repository can be requested at https://www.hmhospitales.com/coronavirus/covid-data-save-lives/english-version

**Supplementary Table 6. Code and other resources made available to the community from work presented in this paper.**

| Folder | Notebook | Content |
|---|---|---|
| 1.cxr_wrangling | hoboken_cxr_features_model.ipynb | In this notebook 320x320 CXR jpg and labels are read from Hoboken CXRs. The model previously trained to classify mortality on the Madrid CXRs is used to extract 64 features and output the prediction from the Hoboken CXR.<br><br>This code is written following a federated approach so only users with Hoboken credentials interact with the CXrs and only needs the *.jpg(s) to be located in their Google Drive. |
| | seoul_cxr_features_model.ipynb | In this notebook 320x320 CXR jpg and labels are read from Seoul CXRs. The model previously trained to classify mortality on the Madrid CXR is used to extract 64 features and output the prediction from the Seoul CXR. |
| | madrid_cxr_augmentation_w_bboxes.ipynb | In this notebook CXR augmentation with bboxes scratch is created, both image and tabular data are loaded in synch with each other, finally iterable loaders that can have the output X1, X2, y for enumerate in order to locate the bbox trachea and lung/heart coordinates. |
| | useful_dicom_metadata.ipynb | In this notebook useful function examples from the pydicom library for exporting selected DICOM metadata into a txt, counting by group and basic QC tests after data conversion from DICOM to jpg. |
| 2.ehr_data_wrangling | madrid_ehr.ipynb | In this notebook Madrid tables are ingested; data is explored; variables are renamed or created; exclusion criteria is applied; vitals, comorbidities, drugs and labs are appropriately transformed and cleaned; tables are joined; table 1 is produced; feature distribution is evaluated; dataset is split into training and internal validation; model is trained with ehr data, evaluated using 4 folds cross-validation and calibrated; model is externally evaluated with other datasets and feature importance is addressed. |
| | hoboken_ehr.ipynb | In this notebook we are preparing the Hoboken test dataset: Hoboken tables are ingested; exclusion criteria is applied; data is explored; vitals, comorbidities, drugs and labs are appropriately transformed and cleaned; variables are mean centered and standardized; missing values are imputed; table 1 is produced; image features are appended and renamed and tables are joined. |
| | seoul_ehr.ipynb | In this notebook we are preparing the Seoul test dataset: Seoul tables are ingested; exclusion criteria are applied; data is explored; vitals, comorbidities, drugs and labs are appropriately transformed and cleaned; variables are |

| | | mean centered and standardized; missing values are imputed; table 1 is produced; image features are appended and renamed and tables are joined. |
|---|---|---|
| 3.training | transfer_from_mimic_plus_model_fusion_tuned.ipynb | In this notebook GCP TPU(s) are loaded; Madrid's previously cleaned training and test CXR jpg(s) and EHR data are loaded; then we put images, EHR data and the labels in the same tf records; EHR data only model is defined; CXR only model is defined; pretrained MIMIC CXR model to identify CheXpert 14 labels is loaded; intermediated fusion model is defined; models are tuned; experiments are set up in order to find the best parameters; CXR model is retrained on full dataset using best parameters; finally results are visualized. |
| | madrid_ehr_and_fusion_4fold.ipynb | In this notebook the code is used to train the EHR-based model, CXR-based model fusion model inMadrid data. Hyper-parameters were tuned using a 4-fold cross validation approach. The results of 4-fold internal validation were also computed. |
| 4.testing | hoboken_test_3models.ipynb | In this notebook the code is used to validate the pretrained EHR-based model, CXR-based model fusion model in the Hoboken dataset. |
| | seoul_test_3models.ipynb | In this notebook the code is used to validate the pretrained EHR-based model, CXR-based model fusion model in the Seoul dataset. |